\documentclass[10pt,twocolumn,letterpaper,x11names]{article}

\usepackage{iccv}              

\definecolor{iccvblue}{rgb}{0.21,0.49,0.74}
\usepackage[pagebackref,breaklinks,colorlinks,allcolors=iccvblue]{hyperref}
\usepackage{mathrsfs}
\usepackage{amsmath}
\usepackage{color}
\usepackage{xcolor}
\usepackage{soul}
\usepackage{colortbl}
\usepackage{diagbox} 

\usepackage{verbatim}
\usepackage{multirow}

\usepackage{float}

\DeclareMathOperator*{\argmax}{arg\,max} 
\DeclareMathOperator*{\argmin}{arg\,min} 
\newcommand{\yxnote}[1]{\textcolor{blue}{#1}}

\def\paperID{2697} 
\def\confName{ICCV}
\def\confYear{2025}

\title{SSAM: Self-Supervised Association Modeling for Test-Time Adaption}

\author{
    \textbf{Yaxiong Wang*\textsuperscript{1}}, \quad
        \textbf{Zhenqiang Zhang*\textsuperscript{1}}, \quad
    \textbf{Lechao Cheng\textsuperscript{1}},\quad
    \textbf{Zhun Zhong\textsuperscript{1}},\\
    \textbf{Dan Guo\textsuperscript{1,2}},\quad
    \textbf{Meng Wang\textsuperscript{1,2}}\\
    \textsuperscript{1}School of Computer Science and Information Engineering, Hefei University of Technology, China\\
    \textsuperscript{2}Institute of Artificial Intelligence, Hefei Comprehensive National Science Center, China\quad 
}

\begin{document}
\maketitle

\begin{abstract}

Test-time adaption (TTA) has witnessed important progress in recent years, the prevailing methods typically first encode the image and the text and design strategies to model the association between them. Meanwhile, the image encoder is usually frozen due to the absence of explicit supervision in TTA scenarios. 
We identify a critical limitation in this paradigm: While test-time images often exhibit distribution shifts from training data, existing methods persistently freeze the image encoder due to the absence of explicit supervision during adaptation. This practice overlooks the image encoder's crucial role in bridging distribution shift between training and test. To address this challenge, we propose SSAM (Self-Supervised Association Modeling), a new TTA framework that enables dynamic encoder refinement through dual-phase association learning. Our method operates via two synergistic components: 1) Soft Prototype Estimation (SPE), which estimates probabilistic category associations to guide feature space reorganization, and 2) Prototype-anchored Image Reconstruction (PIR), enforcing encoder stability through cluster-conditional image feature reconstruction. Comprehensive experiments across diverse baseline methods and benchmarks demonstrate that SSAM can surpass state-of-the-art TTA baselines by a clear margin while maintaining computational efficiency. The framework's architecture-agnostic design and minimal hyperparameter dependence further enhance its practical applicability.



\end{abstract}

\section{Introduction}

Test-time adaptation (TTA) has emerged as a critical paradigm for enhancing model performance on unlabeled test distributions by dynamically adjusting trained models during inference. The fundamental challenge lies in effectively bridging the distribution shift between training and test data, such that the model can perform well on the unseen test samples. Significant research efforts have been devoted to advancing this field~\cite{zhou2022learning,zhou2022conditional,zhang2022tip,shu2022test,feng2023diverse,karmanov2024efficient,zhang2024dual,zhang2024dual2}, with recent progress being particularly accelerated by breakthroughs in vision-language models.

\begin{figure}[t!]
\begin{center}
\includegraphics[width=1\linewidth]{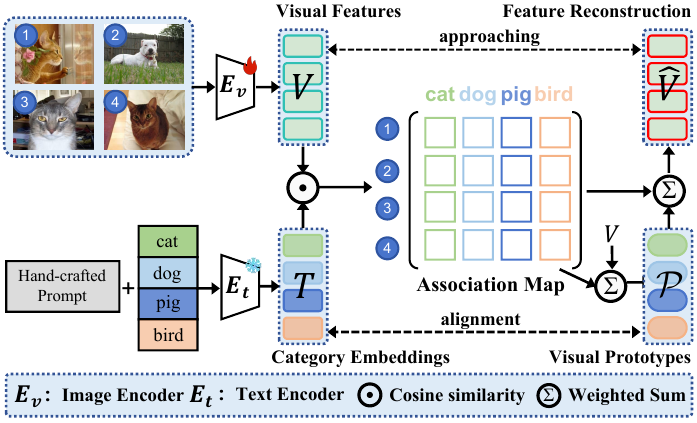}
\end{center}
\vspace{-0.4cm}
\caption{Image instances are fed into a visual encoder to generate the visual feature space. In Soft Prototype Estimation~(SPE), 
an association map is computed by measuring the similarity between visual features and category embeddings.
This association map is then combined with visual features to generate visual prototypes.
In Prototype-anchored Image Reconstruction ~(PIR), visual features are reconstructed using the prototypes and the association map, enabling label-free supervision.}
\label{fig:cscacir}
\end{figure}

The advent of CLIP~\cite{radford2021learning} has revolutionized this landscape through its impressive ability to learn aligned image-text representations from massive data. 
CLIP's zero-shot generalization capabilities have made it particularly amenable to TTA scenarios, where no target data is available~\cite{shu2022test,feng2023diverse}. In recent years, tremendous attempts upon CLIP have been made to address TTA challenges ~\cite{shu2022test,feng2023diverse,karmanov2024efficient,zhang2024dual,zhang2024dual2}.  Roughly, current CLIP-based TTA approaches primarily adopt two  strategies: text prompt tuning and cache-based adaptation. Text prompt methods~\cite{shu2022test,feng2023diverse} optimize learnable vectors in the text embedding space to better capture target domain semantics, with notable examples like TPT introducing specialized loss functions for prompt refinement~\cite{shu2022test}. Cache-based approaches~\cite{zhang2022tip,karmanov2024efficient,zhang2024dual} address data distribution shifts by maintaining memory banks of test features, exemplified by DMN's dual static-dynamic caching mechanism with trainable readout modules~\cite{zhang2024dual}.

Despite these advancements, a critical gap persists in current methodologies. While existing approaches predominantly focus on text encoder adaptation and feature caching, they largely overlook the fundamental role of image encoding in distribution alignment. The inherent category consistency between training and test phases in TTA scenarios suggests that text encoders—operating on fixed semantic concepts—exhibit limited sensitivity to test data variations. By contrast, image encoders must contend with substantial distribution shifts arising from domain discrepancies, style variations, and unforeseen perturbations. This observation compels us to re-examine the prevailing paradigm: effective TTA solutions should prioritize image encoder adaptation as the primary mechanism for test-time distribution alignment.

Motivated by this insight, we propose a Self-Supervised Association Modeling (SSAM) framework for image encoder adaptation. SSAM introduces a lightweight adapter module that dynamically adjusts the image encoding space while preserving frozen backbone parameters. The SSAM architecture operates through two synergistic components: (1) Soft Prototype Estimation (SPE) establishes probabilistic associations between image features and category embeddings, generating prototype vectors that encode distribution characteristics; (2) Prototype-anchored Image Reconstruction  (PIR) leverages these prototypes to reconstruct original features through association-guided weighted combinations. This dual mechanism creates a self-supervised feedback loop where reconstruction fidelity directly reflects accuracy of the association between images and categories, enabling unsupervised adapter optimization. By decoupling adaptation from backbone parameters, SSAM achieves dynamic test-time adjustment while maintaining model stability and preserving original knowledge.

Extensive results on diverse benchmarks and backbones reveal that SSAM is an effective training strategy for TTA problem. In summary, our main contributions are as highlighted as follows:
\begin{itemize}
\item \textbf{Identifying image encoder adaptability as the critical factor in test-time adaption.} Motivated by systematic analysis of distribution shift characteristics, we introduce a new self-supervised association modeling framework that establishes category-aware visual alignment without label supervision.

\item \textbf{A dual-phase optimization paradigm for image encoder adaptation.} Building on this foundation, we develop Soft Prototype Estimation (SPE) and Prototype-anchored Image Reconstruction  (PIR) - two synergistic mechanisms that work in tandem to enable stable and effective image encoder tuning through self-supervised learning.

\item \textbf{Architecture-agnostic framework with empirical superiority.} The proposed SSAM demonstrates remarkable compatibility across diverse backbones while achieving consistent performance improvements, setting new state-of-the-art benchmarks through comprehensive experimental validation.
\end{itemize}

\section{Related Work}
\noindent\textbf{Vision-Language Models (VLMs)} are trained on large-scale image-text pairs collected from the web, enabling them to learn rich semantic representations and align vision and language modalities effectively. Notable VLMs, such as CLIP~\cite{radford2021learning}, leverage contrastive learning to establish strong correlations between images and textual descriptions. By pretraining on diverse datasets and fine-tuning on specific tasks, these models have achieved remarkable performance in various vision tasks, including image recognition~\cite{hegde2023clip,liu2024remoteclip}, object detection~\cite{wu2023cora,wei2023improving}, and depth estimation~\cite{zeng2024wordepth,hu2024learning}. However, adapting VLMs to downstream tasks with limited labeled data remains a challenge. To address this issue, two primary adaptation strategies have been proposed: text prompt tuning~\cite{cho2023distribution,khattak2023maple,shen2024multitask,zhou2022conditional,zhou2022learning,zhu2023prompt} and cache-based adaptation~\cite{gao2024clip,li2024graphadapter,yu2023task,zhang2024negative,zhang2022tip}. These approaches aim to enhance task-specific performance while leveraging the knowledge embedded in pretrained models.

\noindent\textbf{Text prompt tuning} is an effective method for adapting VLMs to downstream tasks~\cite{kenton2019bert,chen2020simple,radford2018improving,radford2018improving} by optimizing learnable vectors in the text embedding space to better capture target domain semantics. Early approaches, such as CoOp~\cite{zhou2022learning}, optimize text prompts by representing context words within the prompts as learnable vectors, which are fine-tuned using labeled data from downstream tasks.
Building upon CoOp~\cite{zhou2022learning}, CoCoOp~\cite{zhou2022conditional} introduces input-conditioned tokens that are generated for each input image. This approach addresses a key limitation of CoOp—its tendency to overfit to base classes seen during training—thus improving generalization to unseen categories. However, both CoOp and CoCoOp rely heavily on labeled data, limiting their applicability in scenarios where annotated samples are scarce.
To mitigate this limitation, TPT~\cite{shu2022test} proposes a test-time prompt tuning strategy that replaces the label-needed cross-entropy with entropy minimization for prompt optimization. By generating augmented views of test images, filtering out low-confidence predictions, and enforcing consistency across augmentations, TPT enhances classification accuracy in scenarios with limited data and sparse annotations.
Building on TPT, DiffTPT~\cite{feng2023diverse} integrates diffusion models for data augmentation, generating diverse visual variations of input images. Furthermore, it employs cosine similarity to filter redundant augmentations, further improving classification performance.

\noindent\textbf{Cache-based adaptation}  methods represent a key approach for adapting VLMs to downstream tasks~\cite{li2022dual,oh2019video,li2023mdqe,karunaratne2021robust} by leveraging downstream data samples to construct cache models tailored for specific applications. Early methods, such as Tip-Adapter~\cite{zhang2022tip}, create cache models by selecting a small subset of representative samples for each category in a classification task. This training-free adapter approach directly constructs both the adapter and cache models, achieving performance comparable to or even surpassing training-based methods. However, Tip-Adapter depends on labeled downstream data for sample selection, limiting its applicability in scenarios with sparse annotations.
To address this limitation, TDA~\cite{karmanov2024efficient} introduces a dynamic adapter that constructs the cache model using high-confidence samples identified during inference. The cache is continuously updated in real time, retaining only the most confident samples while utilizing low-confidence samples to construct a negative cache, which further aids inference by mitigating uncertainty. 
Building upon Tip-Adapter and TDA, DMN~\cite{zhang2024dual} combines the static cache from Tip-Adapter with the dynamic positive cache from TDA. Additionally, DMN incorporates a trainable cache readout function, effectively integrating the advantages of both static and dynamic caches to achieve superior performance.

\section{Preliminaries}
Given a set of test images $\mathcal{I}$ and a trained  model $M$, test-time adaption is to tune $M$ using image set $\mathcal{I}$. Meanwhile,  the category gallery $C=\{C_i\}_{i=1}^{M}$ is is available and remains consistent with the training set, while the specific category of each image belongs to is unknown, ${i.e.,}$ no available labels. 

With the advancement of vision-language models, leveraging the CLIP model as the backbone for TTA has become a dominant strategy, primarily due to its remarkable zero-shot performance in classification tasks. Typically, a modern TTA model comprises two branches: a vision branch and a text branch, both of which are initialized from pre-trained CLIP models. In this setup, the image is first encoded using CLIP's visual encoder to extract a feature representation $V$. Meanwhile, the categories are formatted into text prompts, such as ``\texttt{A Photo of a [CLASS]}'', and passed through the text branch to generate their corresponding feature representations $T$. Finally, the image is classified to the category with the highest similarity to its feature representation $V$.
\begin{equation}
\label{classification}
    C(I) = \argmax_{C_i} \{\text{sim}(I, C_1), \cdots, \text{sim}(I, C_M)\},
\end{equation}
where $C(I)$ is the category of the image $I$ belongs to, $\text{sim}(\cdot,\cdot)$ is a similarity function.

\section{Self-Supervised Association Modeling}
\begin{figure*}[t!]
\begin{center}
\includegraphics[width=1\linewidth]{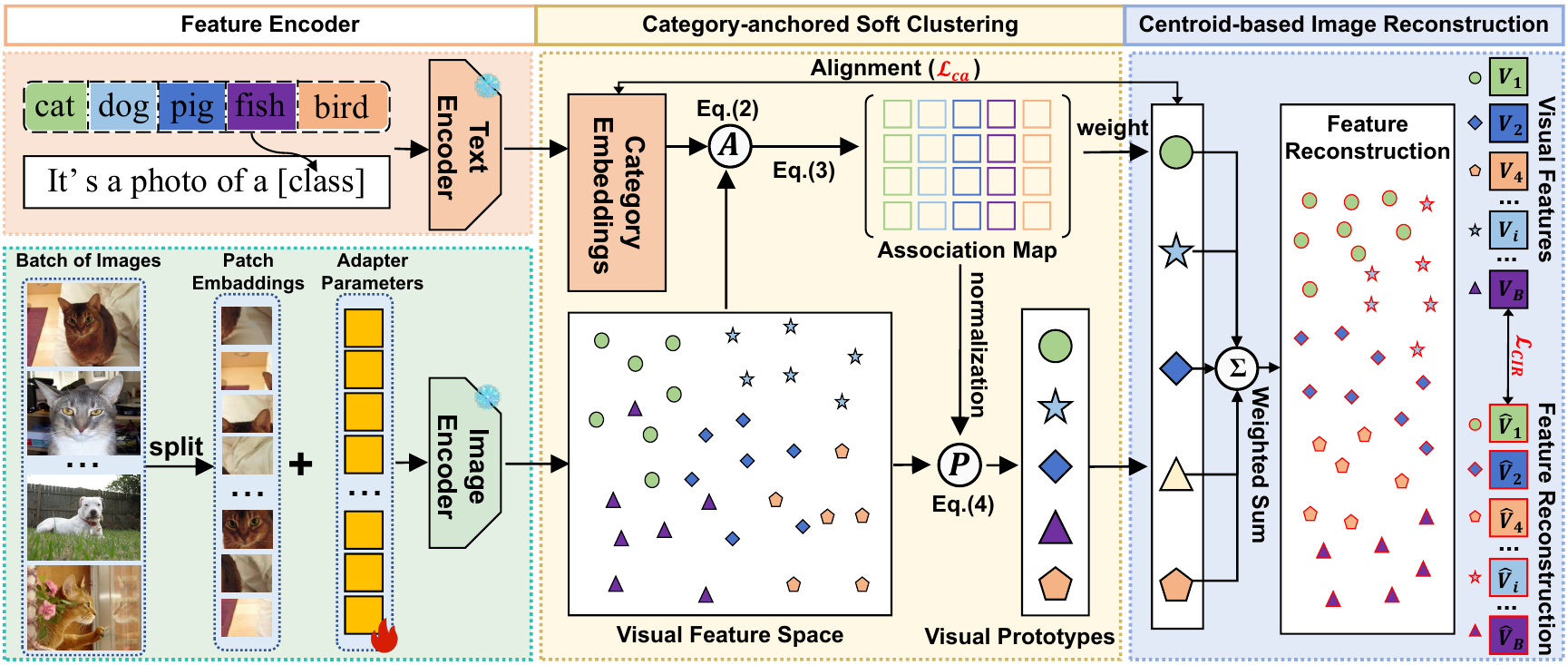}
\end{center}
\vspace{-0.4cm}
\caption{\textbf{Framework of SSAM~(Self-Supervised Association Modeling).} SSAM primarily optimizes the image encoder while preserving the pre-trained parameters of the CLIP visual encoder. 
To achieve this, we introduce a patch-wise image processing strategy, where input images are divided into patches, and lightweight trainable parameters are integrated into the patch encoding process. These parameters are fine-tuned independently, ensuring the original CLIP visual encoder remains unaltered.
SSAM comprises two key components: Soft Prototype Estimation (SPE) and Prototype-anchored Image Reconstruction  (PIR). In CSC, we compute the similarity between extracted image features and category embeddings to construct the association map. The map is then integrated with the image features to generate visual prototypes. In CIR, image reconstruction is performed based on the visual prototypes and the association map. By comparing the reconstructed images with the original inputs, PIR enables label-free fine-tuning of the image encoder, allowing effective adaptation without requiring explicit supervision.}
\label{fig:method}
\end{figure*}
Figure~\ref{fig:method} depicts our Self-Supervised Association Modeling (SSAM) framework. The visual and text encoder processes image and category prompt to generate respective features. Subsequently, in Soft Prototype Estimation~(SPE), we compute the similarity between visual features and category features to construct association map. Then, the map is combined with visual features to generate visual prototypes. In Prototype-anchored Image Reconstruction ~(PIR), we use the prototypes and association map to reconstruct visual features, enabling label-free tuning of the visual encoder.
In the following, we elaborate the details of our CSC and PIR modules, which are the two key designs of our SSAM.

\noindent\textbf{Text Encoder.} We utilize the text branch of the pre-trained CLIP model as our text encoder. The text encoding process begins by tokenizing the text prompt into a sequence of tokens. This token sequence is then appended with a class token and fed into the transformer blocks. Ultimately, the text feature $T$ is derived from the final vectors corresponding to the class token.

\subsection{TTA Adapter for Image Encoder. } 
In this paper, we focus on tuning the image encoder to address the distributional shift between training and test images, as previously discussed. However, tuning the full parameters of the image encoder is neither effective nor efficient. Moreover, the limited number of test samples is insufficient to support full-parameter tuning, which could distort the knowledge embedded in the pre-trained CLIP model. To address these challenges, we propose a lightweight TTA adapter for adaptation while freezing the original parameters of CLIP. Given the significant architectural differences between Vision Transformers and ResNets, the design of the TTA adapter slightly varies accordingly.

\noindent\textbf{TTA adapter for Vision Transformer.} For a test image $I\in \mathcal{I}$, the image is first divided into $N$ non-overlapping patches. Each patch is then flattened and linearly embedded, resulting in a sequence of patch embeddings $\{p_1,p_2,\cdots,p_N\}$, where $p_i\in R^D$.
Subsequently, we introduce $N$ learnable vectors $\phi_a=\{a_1,a_2,\cdots,a_N\}$, each of dimension $D$, as adaptation parameters. These learnable vectors are added to the patch embeddings to produce new patch embeddings as follows:
$q_i=p_i+a_i$.
Finally, the new patch embeddings $\{q_1,q_2,\cdots,q_N\}$ are fed into the subsequent vision attention blocks to obtain the final image feature $V$.

\noindent\textbf{TTA adapter for Resnet. } 
For a test image $I\in \mathcal{I}$, the image is first fed into the first convolutional layer to obtain the hidden representation $O \in \mathbb{R}^{D \times H \times W }$, where $H$ and $W$ are the height and width of the image, respectively. Considering the locality of the image as well as reducing trainable parameters, we adopt a patch-based adaptation: each $s\times s$ patch would share the adaption parameters. Given $N$ trainable tokens, $\phi_a=\{a_1,a_2,\cdots,a_N\}$, each is with $D\times 1\times 1$ dimension and $N=\frac{H}{s} \times \frac{W}{s}$. 
The $i$-th token would be repeated by $s\times s$ times, resulting a new tensor $\bar{a}_i\in \mathbb{R}^{D\times s\times s}$. Subsequently, $\bar{a}_i$ is added to $i$-th patch $o_i$ for adaption: $o_i+\bar{a}_i$. Finally, the newly generated feature maps are passed through the subsequent convolutional layers to produce the final image feature $V$.



\subsection{Soft Prototype Estimation.}

In our Soft Prototype Estimation (SPE), we establish a probabilistic correlation between image features and category embeddings to generate a corresponding probability correlation map. Based on this map, we compute category-specific prototype vectors that capture the distributed characteristics of the data. These prototype vectors serve as guidance for subsequent image reconstruction and correlation matrix optimization, ensuring that the learned representations are both informative and discriminative.

For a batch of images to be tested $B=\{I_1, I_2, \dots, I_{|B|}\}$, we obtain the visual features $\{V_1, V_2, \dots, V_{|B|}\}$ via the image encoder, where  $B$ denotes the mini batch. Meanwhile, the predefined text prompts are encoded by the text encoder, yielding the corresponding features $\{T_1, T_2, \cdots, T_M\}$. Subsequently, we estimate the association between an image and a certain category as follows:
\begin{equation}
\label{cs}
    A(I_i,C_j) = \frac{V_i\times T_j}{||V_i||||T_j||}.
\end{equation}

Next, we normalize the association along the category:
\begin{equation}
\label{ncs}
    \tilde{A}(I_i,C_j) = \frac{\exp{A(I_i,C_j)}}{\sum_{k=1}^M \exp{A(I_i, C_k)}}.
\end{equation}
The resultant $\tilde{A}(I_i,C_j)$ indicates the probability that image $I_i$ belongs the category $C_j$. With this probability, we summary the visual prototype $\mathcal{P}_j $ for each categories as follows:
\begin{equation}
    \mathcal{P}_j = \frac{\sum_{k=1}^{|B|} \tilde{A}(I_k, C_j)\times V_k}{\sum_{k=1}^{|B|} \tilde{A}(I_k, C_j)},
\end{equation}
where the denominator $\sum_{k=1}^{|B|} \tilde{A}(I_k, C_j)$ is the normalization factor.

\subsection{Prototype-anchored Image Reconstruction.}
The association map $\tilde{A}$ is crucial for achieving classification, yet no direct supervision is available for this map. 
To address this challenge, our Prototype-anchored Image Reconstruction (PIR) introduces an indirect yet straightforward supervision approach. Specifically, PIR generates a feature based on the prototype of each category and the probability of the image belonging to those categories:
\begin{equation}
\label{recon}
    {\hat{V_i}} = \sum_{k=1}^{|B|} \tilde{A}(I_i, C_k)\times \mathcal{P}_k.
\end{equation}

From Eq.~\ref{recon}, when the association map successfully predicts an image's categorical membership, the reconstructed feature vector $\hat{V_i}$ should exhibit greater similarity to the original feature set $V_i$, compared to scenarios where the association estimate is less precise. To illustrate, consider an image depicting a lion. If the estimated association map $\tilde{A}$ allocates a higher probability to the lion category, then the resultant vector $\hat{V}_i$ is expected to be more congruent with the actual features $V_i$ than a situation where the association map mistakenly assigns a stronger connection to the cat category. Inspired by this observation, our approach aims to drive the generated vector $\hat{V_i}$ toward a closer approximation of the image's authentic feature representation, $V_i$.
\begin{equation}
\label{reconloss}
    \mathcal{L}_{\text{PIR}} = ||\hat{V_i}-V_i||_2^2.
\end{equation}

Additionally, to establish a strong semantic correspondence between the visual prototype and the category prompt, we incorporate a contrastive alignment strategy between these elements. This inclusion strengthens the reliability of the processes described in Eq.~\ref{recon} and Eq.~\ref{reconloss}:
\begin{equation}
\mathcal{L}_{\text{p2c}} = \frac{1}{M}\sum_{i=1}^{i=M}-\log \frac{\exp(s(\mathcal{P}_i,T_i) )}{ {\textstyle \sum_{T_{k}}^{}}\exp(s(\mathcal{P}_i,T_k)) },
\end{equation}
where $s(\cdot,\cdot)$ is the cosine similarity. In analogue, the alignment $\mathcal{L}_{c2p}$ from category prompt to visual prototype can be formulated in a similar fashion. The final contrastive alignment objectives are:
\begin{equation}
    \mathcal{L}_{ca} = \frac{1}{2}(\mathcal{L}_{p2c}+\mathcal{L}_{c2p}). 
\end{equation}

\noindent\textbf{Discussion.} Assume $I_i$ is an image of $h$-th category. Given $\hat{V}_i$ built from the visual prototypes of the categories (Eq.~\ref{recon}) and the association map, the reconstruction loss in Eq.~\ref{reconloss} actually establishes implicit supervision for association learning: By minimizing Eq.~\ref{reconloss}, we essentially enforce $\hat{V}_i$ to be dominated by the visual prototype $\mathcal{P}_h$ ($I_i$'s true visual prototype). This implicitly demands the association map $\tilde{A}$ to allocate higher values to $\tilde{A}(I_i,C_h)$, thereby ensuring that the learned association matrix $A$ accurately captures the relationship between images and their corresponding categories. Therefore, the primary objective of Eq.~\ref{reconloss} is not to achieve perfect feature reconstruction, but rather to create a self-supervised and indirect optimization way that regularizes the association map learning.

\subsection{Training and Inference.}
\noindent\textbf{Training Objectives.} Following previous practices~\cite{shu2022test,feng2023diverse}, we also employ entropy loss to enhance the confidence of image classification predictions:

\begin{equation}
\mathcal{L}_{ent} = -\frac{1}{|B|}\sum_{i=1}^{|B|}\sum_{j=1}^{M}\tilde{A}(I_i,C_j)\log \tilde{A}(I_i,C_j).
\end{equation}

Further including the reconstruction loss of PIR and the cross alignment constrain, our optimization objective for tuning TTA adapter $\phi_a$ can be outlined as follows:
\begin{equation}
\label{eq:to}
    \argmin_{\phi_a} (\mathcal{L}_{ent} + \alpha \mathcal{L}_{PIR} + \beta \mathcal{L}_{ca}),
\end{equation}
where $\alpha$ and $\beta$ are two hyper-parameters to balance different training objectives.

\noindent\textbf{Inference.} At the stage of inference, we begin by leveraging our fine-tuned image encoder and text encoder to embed both images and categories into their corresponding feature representations. Consistent with established conventions, we proceed to estimate the relationship between each image and the pool of candidate categories, relying on the computations detailed in Eq.~\ref{cs}-\ref{ncs}. Ultimately, the classification decision is made according to Eq. \ref{classification}. As part of our experimental validation, we incorporate our SSAM framework within pre-existing backbone architectures to demonstrate its efficacy. In such instances, the inference protocol remains consistent with the baseline methodologies.

\section{Experiments}
\begin{table*}[t!]
\begin{center}
\resizebox{1\linewidth}{!}{
\begin{tabular}{llccccccccccc}
\toprule[1.5pt]
\rowcolor[gray]{.93} Method &Reference          & Aircraft & Caltech101 & Cars  & DTD   & EuroSAT & Flower102 & Food101 & Pets  & SUN397 & UCF101 & Average \\ \hline
\multicolumn{13}{c}{\underline{\textbf{\quad Backbone: ResNet-50 \quad}}} \\

CoOp~\cite{zhou2022learning}     &IJCV'22  & 15.12    & 86.53      & 55.32 & 37.29 & 26.20   & 61.55     & 75.59   & 87.00 & 58.15  & 59.05  & 56.18   \\
CoCoOp~\cite{zhou2022conditional}&CVPR'22  & 14.61    & 87.38      & 56.22 & 38.53 & 28.73   & 65.57     & 76.20   & 88.39 & 59.61  & 57.10  & 57.23   \\ 
TPT~\cite{shu2022test}           &NIPS'22  & 17.58    & 87.02      & 58.46 & 40.84 & 28.33   & 62.69     & 74.88   & 84.49 & 61.46  & 60.82  & 57.66   \\
DiffTPT~\cite{feng2023diverse}   &ICCV'23  & 17.60    & 86.89      & 60.71 & 40.72 & 41.04   & 63.53     & 79.21   & 83.40 & 62.72  & 62.67  & 59.85   \\ \hline
CLIP~\cite{radford2021learning}  &ICML'21  & 16.11    & 87.26      & 55.89 & 40.37 & 25.79   & 62.77     & 74.82   & 82.97 & 60.85  & 59.48  & 56.63   \\
\rowcolor[gray]{.95}$\text{SSAM}_\text{CLIP}$    &OURS & \textbf{17.22}    & \textbf{88.07}      & \textbf{56.66} & \textbf{42.38} & \textbf{47.88}   & \textbf{66.18}     & \textbf{77.38}   & \textbf{85.96} & \textbf{61.21}  & \textbf{61.86}  & \hspace{-0.1cm}$\text{\qquad\textbf{60.48}}_{\textbf{3.85}\uparrow}$   \\
TDA~\cite{karmanov2024efficient} &CVPR'24  & \textbf{17.61}    & 89.70      & \textbf{57.78} & 43.74 & 42.11   & 68.74     & \textbf{77.75}   & 86.18 & \textbf{62.53}  & 64.18  & 61.03   \\ 
\rowcolor[gray]{.95}$\text{SSAM}_\text{TDA}$     &OURS & 17.49    & \textbf{89.74}      & 57.73 & \textbf{44.86} & \textbf{54.94}   & \textbf{68.90}     & 77.68   & \textbf{86.24} & 62.51  & \textbf{64.42}  & \hspace{-0.1cm}$\text{\qquad\textbf{62.45}}_{\textbf{1.42}\uparrow}$   \\
DPE~\cite{zhang2024dual2}        &NIPS'24  & 19.80    & 90.83      & 59.26 & 50.18 & 41.67   & 67.60     & 77.83   & 85.97 & 64.23  & 61.98  & 61.93   \\
\rowcolor[gray]{.95}$\text{SSAM}_\text{DPE}$     &OURS & \textbf{21.75}    & \textbf{90.83}      & \textbf{59.68} & \textbf{54.08} & \textbf{56.14}   & \textbf{69.10}     & 77.46   & \textbf{86.73} & \textbf{64.72}  & \textbf{63.84}  & \hspace{-0.1cm}$\text{\qquad\textbf{64.43}}_{\textbf{2.50}\uparrow}$   \\
DMN-ZS~\cite{zhang2024dual}      &CVPR'24  & 22.77    & 90.14      & 60.02 & 50.41 & 48.72   & 67.93     & 76.70   & 86.78 & 64.39  & 65.34  & 63.32   \\
\rowcolor[gray]{.95}$\text{SSAM}_\text{DMN-ZS}$  &OURS & \textbf{22.95}    & \textbf{90.71}      & 59.79 & \textbf{51.00} & \textbf{55.58}   & \textbf{68.82}     & 76.49   & 86.48 & \textbf{64.53}  & 65.13  & \hspace{-0.1cm}$\text{\qquad\textbf{64.15}}_{\textbf{0.83}\uparrow}$   \\ \hline \hline
\multicolumn{13}{c}{\underline{\textbf{\quad Backbone: ViT-B/16 \quad}}} \\
CoOp~\cite{zhou2022learning}     &IJCV'22  & 18.47    & 93.70      & 64.51 & 41.92 & 46.39   & 68.71     & 85.30   & 89.14 & 64.15  & 66.55  & 63.88   \\
CoCoOp~\cite{zhou2022conditional}&CVPR'22  & 22.29    & 93.79      & 64.90 & 45.45 & 39.23   & 70.85     & 83.97   & 90.46 & 66.89  & 68.44  & 64.63   \\ 
TPT~\cite{shu2022test}           &NIPS'22  & 24.78    & 94.16      & 66.87 & 47.75 & 42.44   & 68.98     & 84.67   & 87.79 & 65.50  & 68.04  & 65.10   \\
DiffTPT~\cite{feng2023diverse}   &ICCV'23  & 25.60    & 92.49      & 67.01 & 47.00 & 43.13   & 70.10     & 87.23   & 88.22 & 65.74  & 62.67  & 65.47   \\ \hline
CLIP~\cite{radford2021learning}  &ICML'21  & 23.22    & 93.55      & 66.11 & 45.04 & 50.42   & 66.99     & 82.86   & 86.92 & 65.63  & 65.16  & 64.59   \\
\rowcolor[gray]{.95}$\text{SSAM}_\text{CLIP}$    &OURS & \textbf{24.87}    & \textbf{94.36}      & \textbf{66.80} & 44.74 & \textbf{68.40}   & \textbf{71.78}     & \textbf{85.94}   & \textbf{89.23} & \textbf{66.92}  & \textbf{67.75}  & \hspace{-0.1cm}$\text{\qquad\textbf{68.08}}_{\textbf{3.49}\uparrow}$ \\
TDA~\cite{karmanov2024efficient} &CVPR'24  & 23.91    & 94.24      & 67.28 & 47.40 & 58.00   & 71.42     & 86.14   & 88.63 & 67.62  & 70.66  & 67.53   \\ 
\rowcolor[gray]{.95}$\text{SSAM}_\text{TDA}$     &OURS & \textbf{25.68}    & \textbf{94.77}      & \textbf{68.25} & 46.63 & \textbf{75.37}   & \textbf{72.47}     & \textbf{86.32}   & \textbf{90.32} & \textbf{68.12}  & \textbf{71.19}  & \hspace{-0.1cm}$\text{\qquad\textbf{69.91}}_{\textbf{2.38}\uparrow}$  \\
DPE~\cite{zhang2024dual2}        &NIPS'24  & 28.95    & 94.81      & 67.31 & 54.20 & 55.79   & 75.07     & 86.17   & 91.14 & 70.07  & 70.44  & 69.40   \\
\rowcolor[gray]{.95}$\text{SSAM}_\text{DPE}$     &OURS & \textbf{29.88}    & \textbf{95.62}      & \textbf{69.34} & \textbf{56.26} & \textbf{69.16}   & \textbf{75.51}     & \textbf{86.35}   & \textbf{91.36} & \textbf{70.27}  & \textbf{72.77}  & \hspace{-0.1cm}$\text{\qquad\textbf{71.65}}_{\textbf{2.25}\uparrow}$   \\
DMN-ZS~\cite{zhang2024dual}      &CVPR'24  & 30.03    & 95.38      & 67.96 & 55.85 & 59.43   & 74.49     & 85.08   & 92.04 & 70.18  & 72.51  & 70.30   \\
\rowcolor[gray]{.95}$\text{SSAM}_\text{DMN-ZS}$  &OURS & \textbf{30.81}    &\textbf{95.42}      & \textbf{68.87} & \textbf{55.85} & \textbf{75.86}   & \textbf{75.76}     & 84.85   & 91.99 & \textbf{70.40}  & \textbf{72.85}  & \hspace{-0.1cm}$\text{\qquad\textbf{72.27}}_{\textbf{1.97}\uparrow}$  \\
\bottomrule[1.5pt]
\end{tabular}}
\end{center}
\vspace{-0.4cm}
\caption{\textbf{Results on the Cross-Domain (CD) Benchmark}, where \textbf{bold} indicates the better performance in each comparison pair (baseline \emph{vs.} \colorbox{lightgray!25}{SSAM+baseline}). The CD Benchmark primarily assesses the adaptability of our method to domain shifts across different fields. The evaluation metric \textbf{Average} is computed as the mean accuracy across all ten datasets. }
\label{tab:CDBenchmark}
\end{table*}
\subsection{Experiment Settings}
\textbf{Datasets.} Following the settings in previous works~\cite{shu2022test,feng2023diverse,karmanov2024efficient,zhang2024dual,zhang2024dual2}, we validate the effectiveness of our method on Cross-Domain (CD) Benchmark, including Flowers102~\cite{nilsback2008automated}, DTD~\cite{cimpoi2014describing}, OxfordPets~\cite{parkhi2012cats}, StanfordCars~\cite{krause20133d}, UCF101~\cite{soomro2012ucf101}, Caltech101~\cite{fei2004learning}, Food101~\cite{bossard2014food}, SUN397~\cite{xiao2010sun}, FGVCAircraft~\cite{maji2013fine}, and EuroSAT~\cite{helber2019eurosat}. These datasets cover a wide range of domains and are used to assess the adaptability of our method to domain shifts across different fields.
Additionally, we evaluate our method on Out-of-Distribution (OOD) Benchmark, including ImageNet~\cite{deng2009imagenet}, ImageNet-A~\cite{hendrycks2021natural}, ImageNet-R~\cite{hendrycks2021many}, and ImageNet-Sketch~\cite{wang2019learning}. The latter four datasets are derived from ImageNet and are specifically designed to assess model robustness and generalization to unseen data.

\noindent\textbf{Implementation Details.} We validate our method on classic model CLIP~\cite{radford2021learning} and three state-of-the-art test-time adaptation methods: DMN~\cite{zhang2024dual}, TDA~\cite{karmanov2024efficient}, and DPE~\cite{zhang2024dual2}. For the visual encoder, we employ two widely-used backbones pre-trained with CLIP: ResNet50~\cite{he2016deep} and ViT-B/16~\cite{alexey2020image}. 
To adapt both backbones, we introduce lightweight trainable vectors as adapter parameters, which fine-tune the visual encoder while keeping the backbone frozen.
For DMN, we adopt its train-free zero-shot version for a fair evaluation. 
The initial learning rate is set to 1e-4, and the batch size is set to 256. 
Regarding hyper-parameters $\alpha$ and $\beta$ in Equation~\ref{eq:to}, both are set to 1.0 to balance the three training objectives.
All details will be given in our code repertory.
\subsection{Quantitative Comparison}
In this section, we evaluate the performance of our method on the Cross-Domain (CD) benchmark that covers diverse categories from various domains, and the Out-Of-Distribution (OOD) benchmark includes the ImageNet dataset with four extended datasets.

\noindent\textbf{Evaluation on the Cross-Domain (CD) Benchmark.}
The primary goal of this evaluation is to assess the adaptability of our method to different domain shifts. The detailed results are presented in Table~\ref{tab:CDBenchmark}, where $\text{SSAM}_\text{CLIP}$ means the CLIP model trained with our SSAM. The results clearly demonstrate that our proposed SSAM substantially enhances performance across CLIP~\cite{radford2021learning} and three widely adopted TTA methods~\cite{karmanov2024efficient,zhang2024dual,zhang2024dual2}. 
On the ResNet-50 backbone, our DPE-based method achieves the best performance, surpassing the current state-of-the-art DMN-ZS~\cite{zhang2024dual} by 1.11\% in average performance. Meanwhile, on the ViT-B/16 backbone, our DMN-ZS-based approach yields the highest performance, outperforming DMN-ZS by 1.97\% in average performance. These improvements demonstrate the effectiveness of our method when dealing with cross-domain datasets.

\begin{table*}[t!]
\begin{center}
\resizebox{0.85\linewidth}{!}{
\begin{tabular}{llccccccc}
\toprule[1.5pt]
\rowcolor[gray]{.93} Method &Reference       & ImageNet & ImageNet-A & ImageNet-V2 & ImageNet-R & ImageNet-S & Average & OOD Average \\ \hline
\multicolumn{9}{c}{\underline{\textbf{\quad Backbone: ResNet-50 \quad}}} \\
CoOp~\cite{zhou2022learning}            &IJCV'22  & 63.33    & 23.06      & 55.40       & 56.60      & 34.67      & 46.61    &42.43 \\
CoCoOp~\cite{zhou2022conditional}       &CVPR'22  & 62.81    & 23.32      & 55.72       & 57.74      & 34.48      & 46.81    &42.82 \\
Tip-Adapter~\cite{zhang2022tip}         &ECCV'22  & 62.03    & 23.13      & 53.97       & 60.35      & 35.74      & 47.04    &43.30 \\ 
TPT~\cite{shu2022test}                  &NIPS'22  & 60.74    & 26.67      & 54.70       & 59.11      & 35.09      & 47.26    &43.89 \\
DiffTPT~\cite{feng2023diverse}          &ICCV'23  & 60.80    & 31.06      & 55.80       & 58.80      & 37.10      & 48.71    &45.69 \\ \hline
CLIP~\cite{radford2021learning}         &ICML'21  & 59.81    & 23.24      & 52.91       & 60.72      & 35.48      & 46.43    &43.09 \\ 
\rowcolor[gray]{.95}$\text{SSAM}_\text{CLIP}$  &OURS & \textbf{60.32}    & \textbf{30.25}      & \textbf{55.00}       & \textbf{62.37}      & \textbf{36.24}      & \hspace{-0.1cm}$\text{\qquad\textbf{48.84}}_{\textbf{2.41}\uparrow}$    &\hspace{-0.1cm}$\text{\qquad\textbf{45.97}}_{\textbf{2.88}\uparrow}$ \\ 
TDA~\cite{karmanov2024efficient}        &CVPR'24  & \textbf{61.35}    & 30.29      & \textbf{55.54}       & 62.58      & 38.12      & 49.58    &46.63\\  
\rowcolor[gray]{.95}$\text{SSAM}_\text{TDA}$   &OURS  & 61.22    & \textbf{30.71}      & 55.39       & \textbf{62.95}      & \textbf{38.22}      & \hspace{-0.1cm}$\text{\qquad\textbf{49.70}}_{\textbf{0.12}\uparrow}$    & \hspace{-0.1cm}$\text{\qquad\textbf{46.82}}_{\textbf{0.19}\uparrow}$\\
DMN-ZS~\cite{zhang2024dual}             &CVPR'24  & 63.87    & 28.57      & 56.12       & 61.44      & 39.84      & 49.97    &46.49 \\
\rowcolor[gray]{.95}$\text{SSAM}_\text{DMN-ZS}$&OURS  & \textbf{63.99}    & \textbf{28.67}      & \textbf{56.61}       & \textbf{61.50}      & \textbf{40.31}      & \hspace{-0.1cm}$\text{\qquad\textbf{50.22}}_{\textbf{0.25}\uparrow}$    &\hspace{-0.1cm}$\text{\qquad\textbf{46.77}}_{\textbf{0.28}\uparrow}$\\ 
DPE~\cite{zhang2024dual2}               &NIPS'24  & 63.41    & \textbf{30.15}      & 56.72       & \textbf{63.72}      & 40.03      & 50.81    &47.66 \\
\rowcolor[gray]{.95}$\text{SSAM}_\text{DPE}$   &OURS  & \textbf{63.77}    & 29.99      & \textbf{57.09}       & 63.64      & \textbf{40.33}      & \hspace{-0.1cm}$\text{\qquad\textbf{50.96}}_{\textbf{0.15}\uparrow}$    &\hspace{-0.1cm}$\text{\qquad\textbf{47.76}}_{\textbf{0.10}\uparrow}$ \\ \hline \hline
\multicolumn{9}{c}{\underline{\textbf{\quad Backbone: ViT-B/16 \quad}}} \\
CoOp~\cite{zhou2022learning}            &IJCV'22  & 71.51    & 49.71      & 64.20       & 75.21      & 47.99      & 61.72    &59.28 \\
CoCoOp~\cite{zhou2022conditional}       &CVPR'22  & 71.02    & 50.63      & 64.07       & 76.18      & 48.75      & 62.13    &59.91 \\
Tip-Adapter~\cite{zhang2022tip}         &ECCV'22  & 70.75    & 51.04      & 63.41       & 77.76      & 48.88      & 62.37    &60.27 \\ 
TPT~\cite{shu2022test}                  &NIPS'22  & 68.98    & 54.77      & 63.45       & 77.06      & 47.94      & 62.44    &60.81 \\
DiffTPT~\cite{feng2023diverse}          &ICCV'23  & 70.30    & 55.68      & 65.10       & 75.00      & 46.80      & 62.28    &60.52 \\ \hline
CLIP~\cite{radford2021learning}         &ICML'21  & 68.34    & 49.89      & 61.88       & 77.65      & 48.24      & 61.20    &59.42 \\
\rowcolor[gray]{.95} $\text{SSAM}_\text{CLIP}$ &OURS  & \textbf{69.29}    & \textbf{60.58}     &\textbf{64.79}  & \textbf{80.23}  & \textbf{49.70}      &\hspace{-0.1cm}$\text{\qquad\textbf{64.92}}_{\textbf{3.72}\uparrow}$    &\hspace{-0.1cm}$\text{\qquad\textbf{63.83}}_{\textbf{4.41}\uparrow}$ \\ 
TDA~\cite{karmanov2024efficient}        &CVPR'24  & 69.51    & 60.11      & 64.67       & 80.24      & 50.54      & 65.01   & 63.89\\   
\rowcolor[gray]{.95}$\text{SSAM}_\text{TDA}$    &OURS  & \textbf{69.83}    & \textbf{60.83}      & \textbf{64.97}       & \textbf{80.59}      & \textbf{50.95}      & \hspace{-0.1cm}$\text{\qquad\textbf{65.43}}_{\textbf{0.42}\uparrow}$   & \hspace{-0.1cm}$\text{\qquad\textbf{64.34}}_{\textbf{0.45}\uparrow}$\\
DMN-ZS~\cite{zhang2024dual}             &CVPR'24  & 72.25    & 58.28      & 65.17       & 78.55      & 53.20      & 65.49    &63.80 \\
\rowcolor[gray]{.95}$\text{SSAM}_\text{DMN-ZS}$ &OURS  & \textbf{72.46}    & \textbf{59.24}      & \textbf{65.45}       & \textbf{78.75}      & \textbf{53.79}      & \hspace{-0.1cm}$\text{\qquad\textbf{65.94}}_{\textbf{0.45}\uparrow}$    &\hspace{-0.1cm}$\text{\qquad\textbf{64.31}}_{\textbf{0.51}\uparrow}$ \\ 
DPE~\cite{zhang2024dual2}               &NIPS'24  & 71.91    & 59.63      & 65.44       & \textbf{80.40}      & 52.26      & 65.93    &64.43 \\
\rowcolor[gray]{.95}$\text{SSAM}_\text{DPE}$    &OURS  & \textbf{72.11}    & \textbf{60.71}      & \textbf{65.50}       & 80.39      & \textbf{52.60}      & \hspace{-0.1cm}$\text{\qquad\textbf{66.26}}_{\textbf{0.33}\uparrow}$    &\hspace{-0.1cm}$\text{\qquad\textbf{64.80}}_{\textbf{0.37}\uparrow}$ \\ 
\bottomrule[1.5pt]
\end{tabular}}
\end{center}
\vspace{-0.4cm}
\caption{\textbf{Results on the Out-of-Distribution (OOD) Benchmark.} The OOD Benchmark primarily assesses the robustness and generalization capability of methods when applied to unseen data. The two evaluation metrics, \textbf{Average} and \textbf{OOD Average}, are computed as the mean accuracy across all five datasets and the four OOD datasets, respectively, with ImageNet excluded from the latter.}
\label{tab:OODBenchmark}
\end{table*}

\noindent\textbf{Evaluation on the Out-of-Distribution (OOD) Benchmark.} 
The primary objective of this evaluation is to assess SSAM's robustness and generalization to unseen data. The detailed experimental results are presented in Table~\ref{tab:OODBenchmark}. The results demonstrate that our method consistently enhances performance across CLIP~\cite{radford2021learning} and the three TTA methods~\cite{karmanov2024efficient,zhang2024dual,zhang2024dual2}. Notably, CLIP exhibits the most significant improvement, with average performance gains of 2.41\% and 3.72\% on the ResNet-50 and ViT-B/16 backbones, respectively. Our DPE-based approach achieves average performances of 50.96\% on the ResNet-50 and 66.26\% on the ViT-B/16 backbones, establishing a new state-of-the-art compared to existing TTA methods. These results emphasize the robustness and excellent generalization ability of our method in handling unseen data.

\subsection{Ablation and Visualization}
\begin{figure*}[t!]
\begin{center}
\includegraphics[width=0.8\linewidth]{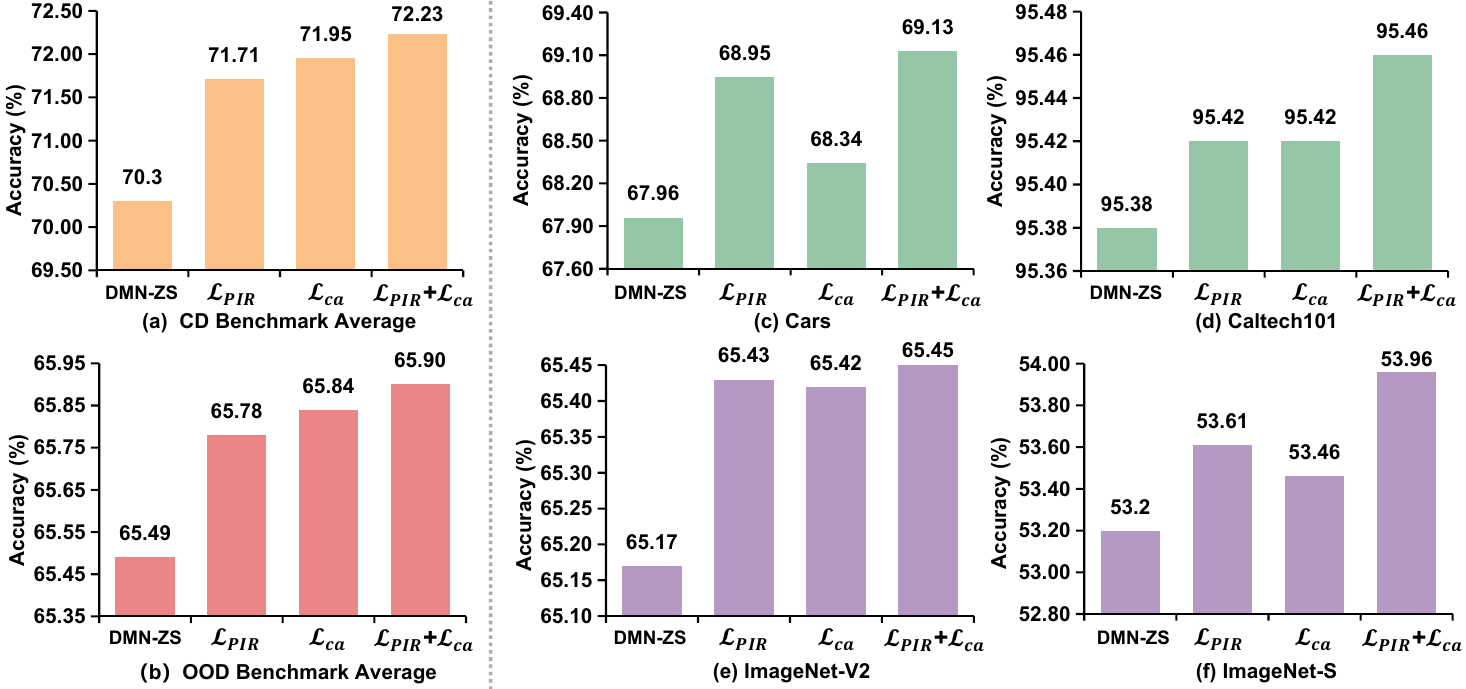}
\end{center}
\vspace{-0.55cm}
\caption{We conduct experiments based on the DMN-ZS model~\cite{zhang2024dual} to assess effectiveness of $\mathcal{L}_{CIR}$ and $\mathcal{L}_{ca}$.
Figure (a) shows the average performance across the CD benchmark, while Figures (c) and (d) depict the performance of two typical datasets in the CD benchmark.
Figure~(b) presents the average performance across the OOD benchmark, including ImageNet. Figures (e) and (f) illustrate the performance of two representative datasets in the OOD benchmark. 
All experiments are based on the VIT-B/16 backbone.
}
\label{fig:effectivenessOfLoss}
\end{figure*}

\textbf{Ablation study on the effectiveness of $\mathcal{L}_{CIR}$ and $\mathcal{L}_{ca}$.}
We conduct an ablation study to evaluate the impact of $\mathcal{L}_{CIR}$ and $\mathcal{L}_{ca}$ on model performance, both individually and in combination. 
The detailed experimental results, presented in Figure~\ref{fig:effectivenessOfLoss}, demonstrate that applying either the alignment loss or the reconstruction loss individually enhances model performance. Furthermore, their combined application leads to more significant performance improvements. After applying the both losses, our method achieves average performances of 72.23\% and 65.90\% on the CD and OOD benchmarks, respectively. This leads to improvements of 1.93\% and 0.41\% compared to DMN-ZS~\cite{zhang2024dual}.
The performance improvement in Figure~\ref{fig:effectivenessOfLoss} validates the effectiveness of two loss functions

\begin{figure*}[t!]
\begin{center}
\includegraphics[width=1\linewidth]{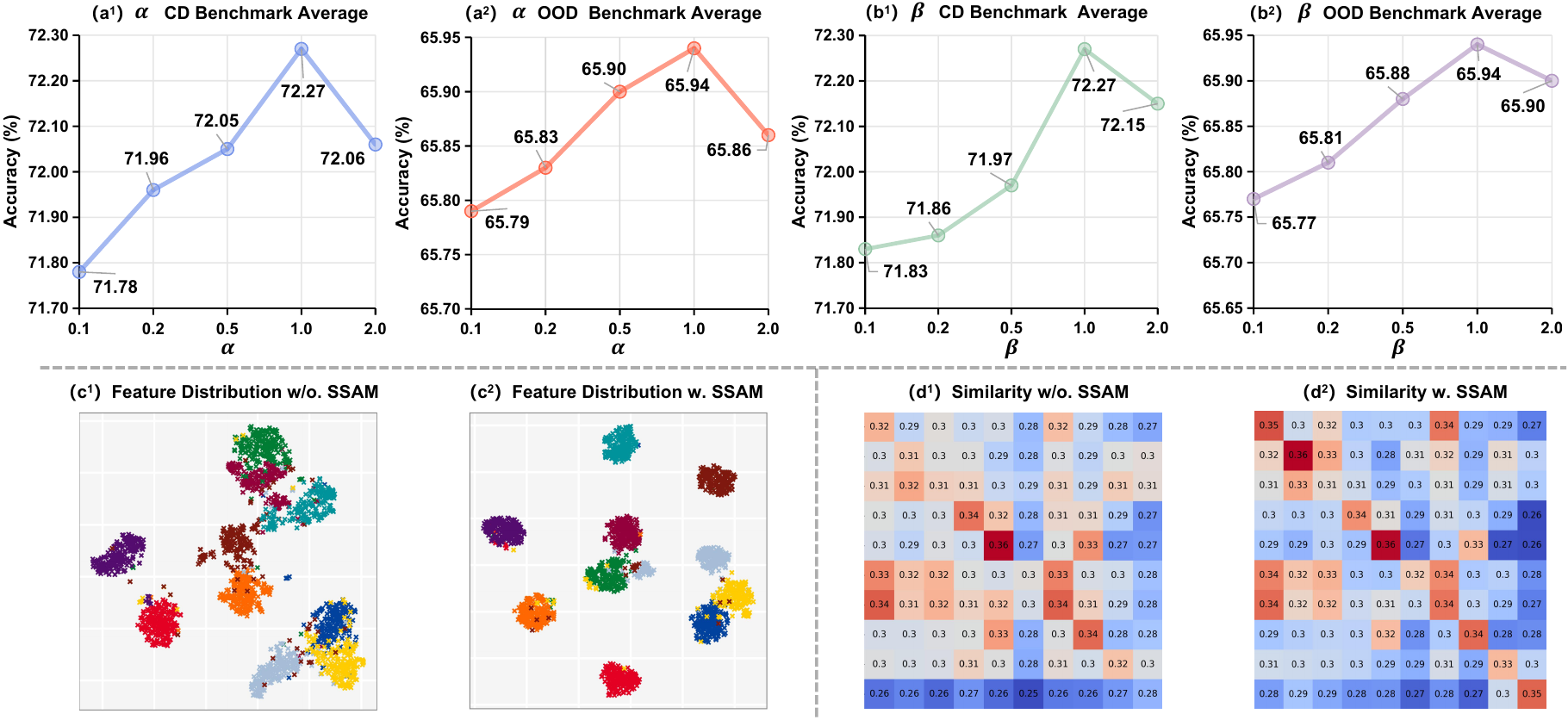}
\end{center}
\vspace{-0.55cm}
\caption{Figure (a) presents the analysis of the impact of parameter $\alpha$ under a fixed $\beta$, 
where \textbf{CD Benchmark Average} represents the average performance across ten cross-domain datasets,
and \textbf{OOD Benchmark Average} represents the average performance across five datasets, including ImageNet. 
Similarly, Figure (b) illustrates the analysis of parameter $\beta$ under a fixed $\alpha$. 
Figure (c) shows the change in visual feature distribution of the EuroSAT dataset before and after applying SSAM. 
Figure (d) depicts the similarity between visual features and textual category embeddings in the EuroSAT dataset before and after applying SSAM, where the horizontal axis represents the textual category features, and the vertical axis represents the visual features.
All experiments are based on the VIT-B/16 backbone.
}
\label{fig:alpha8beta8tsne8heatmap}
\end{figure*}

\noindent\textbf{Ablation study on the influence of hyper-parameters $\alpha$ and $\beta$.}
We conduct experiments to evaluate the impact of the training objective balance hyper-parameters $\alpha$ and $\beta$ in Eq.~\ref{eq:to} on the performance of SSAM. In these experiments, we first fix one parameter while setting the other parameter to 0.1, 0.2, 0.5, 1.0, and 2.0. The detailed results are presented in Figure~\ref{fig:alpha8beta8tsne8heatmap}~(a) and (b). 
As shown in the figure, when both $\alpha$ and $\beta$ are set to 1.0, the model achieves the best performance, with average performances of 72.23\% and 65.90\% on the CD and OOD benchmarks, respectively.
When $\alpha$ is small, the model prioritizes aligning textual and visual features over image reconstruction in the CIR. In contrast, a small $\beta$ leads to excessive focus on image reconstruction, weakening the impact of $\mathcal{L}_{ca}$ and disrupting visual-textual features alignment.

\noindent\textbf{Visualization of features distribution.}
We perform a visual analysis of the feature distribution on the EuroSAT~\cite{helber2019eurosat} dataset using a visual encoder based on the ViT-B/16 backbone. The EuroSAT dataset is selected due to its smaller number of categories, which facilitates clearer visualizations. The detailed results are shown in Figure~\ref{fig:alpha8beta8tsne8heatmap}~(c). By comparing Figures~\ref{fig:alpha8beta8tsne8heatmap}~($\text{c}^1$) and~($\text{c}^2$), before and after applying SSAM, we observe that SSAM leads to a more dispersed distribution of visual features across categories. This increased feature dispersion contributes to improved performance in image classification tasks. The experimental results in the figure validate the effectiveness of our method.

\noindent\textbf{Visualization of similarity between visual and textual features.}
We visualize the similarity between visual and textual features in the EuroSAT~\cite{helber2019eurosat} dataset, with the detailed results presented in Figure~\ref{fig:alpha8beta8tsne8heatmap}~(d). By comparing the results before and after applying the SSAM, as illustrated separately in Figures~\ref{fig:alpha8beta8tsne8heatmap}~($\text{d}^1$) and~($\text{d}^2$).  we observe a significant increase in similarity along the diagonal after SSAM is applied. This enhancement indicates that our method enhances visual-textual feature alignment, leading to more accurate comparisons during classification and a substantial boost in classification performance. 
These findings further validate the effectiveness of our method.

\section{Conclusion}
In this work, we introduce the Self-Supervised Association Modeling~(SSAM) method, a novel self-supervised approach tailored to address the Test-Time Adaptation (TTA) problem. Unlike conventional methods such as text prompt tuning and cache-based adapters, SSAM emphasizes the optimization of key visual encoders. SSAM comprises two sequential steps: Soft Prototype Estimation (SPE) and Prototype-anchored Image Reconstruction  (PIR). CSC estimates the prototype vectors of probabilistic category associations by integrating visual features with their category probabilities, thereby guiding the reorganization of the feature space. PIR employs cluster-conditional image generation to provide label-free supervision for the visual encoder, thereby improving their adaptability and stability. Extensive experiments on both CD and OOD benchmarks indicate that our proposed SSAM achieves state-of-the-art performance. Our study not only contributes to the research on test-time adaptation but also offers a practical solution for improving the efficiency of TTA in visual language models. Overall, SSAM provides a promising direction for advancing the field of test-time adaptation.

\clearpage
{
    \small
    \bibliographystyle{ieeenat_fullname}
    \bibliography{main}
}
\clearpage

\appendix
\maketitlesupplementary
\pagenumbering{arabic}
\setcounter{page}{1}

\section{Discussion on the influence of the adapter parameters position}

\begin{figure}[t!]
\begin{center}
\includegraphics[width=1\linewidth]{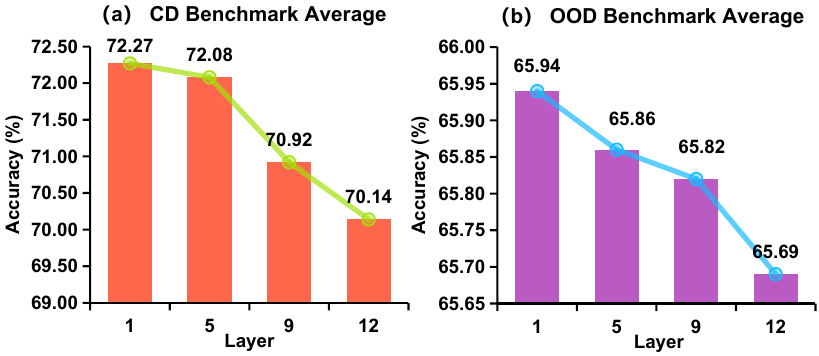}
\end{center}
\vspace{-0.4cm}
\caption{\textbf{Discussion on the influence of the adapter parameters position.} The figures illustrate the variations in average performance across CD and OOD benchmarks as the adapter parameter is positioned differently. 
All experiments are based on the VIT-B/16 backbone.}
\label{fig:sm_vitlayer}
\end{figure}

In our design, for the ViT-B/16 backbone, the adapter parameters are inserted before the first layer of the 12-layer attention blocks. 
To assess the impact of adapter position on model performance, we conduct experiments by positioning the adapter at different layers: before the fifth layer, before the ninth layer, and after the twelfth layer.  
The results, presented in Figure~\ref{fig:sm_vitlayer}, indicate a gradual decline in performance as the adapter is placed in deeper layers. 
The average performance of $\text{SSAM}_\text{DMN-ZS}$ on the CD Benchmark decreased from 72.27\% to 70.14\%, while its performance on the OOD Benchmark dropped from 65.94\% to 65.69\%.
This performance degradation occurs because inserting adapters in higher layers interferes with the extracted high-level feature representations, thereby reducing the model's overall expressiveness. 
In contrast, placing adapters in shallower layers allows the model to better refine low-level feature extraction, supporting a more effective hierarchical learning process.
When adapters are introduced at deeper layers, the model has already constructed stable semantic representations, and modifying these features at this stage disrupts the classification decision boundaries, leading to reduced generalization. Therefore, placing the adapter before the first attention block enhances feature extraction in the early stages, ultimately resulting in improved classification performance.

\section{Visualization of features distribution and visual-textual similarity}
To analyze the impact of the SSAM module on feature distribution, we visualize the changes in the visual feature distribution of the EuroSAT dataset before and after integrating SSAM into the ViT-B/16 backbone. To further validate the effectiveness of SSAM, we extend this visualization experiment to the ResNet-50 backbone, with the corresponding results presented in Figure~\ref{fig:sm_alpha8beta}~(a). The experimental results indicate that the transformation in feature distribution observed on ResNet-50 aligns with the results seen in ViT-B/16.  Specifically, applying SSAM results in a more dispersed feature distribution, suggesting that SSAM effectively enhances feature diversity. This increased dispersion not only improves the model’s ability to generalize across different tasks but also contributes to more accurate image classification by facilitating a broader range of feature representations that are crucial for distinguishing between complex image categories.

\begin{figure}[t!]
\begin{center}
\includegraphics[width=1\linewidth]{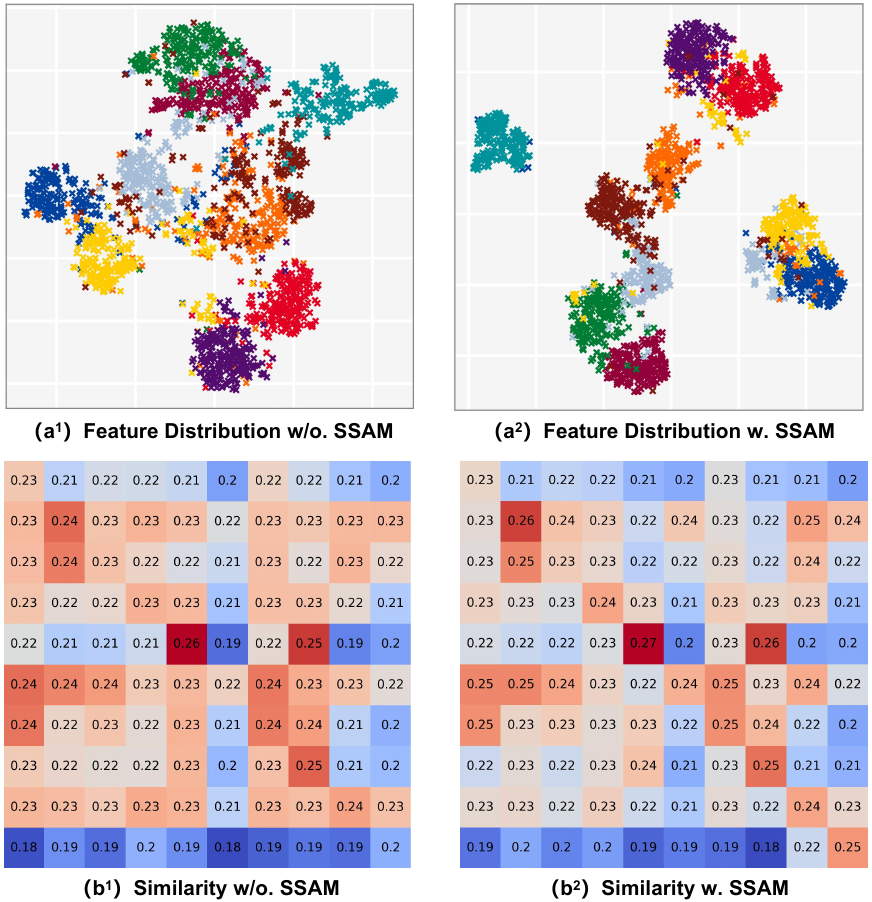}
\end{center}
\vspace{-0.4cm}
\caption{\textbf{Visualization of feature distribution and the similarity between visual and textual features.} Figure (a) illustrate the shifts in visual feature distribution before and after applying SSAM, whereas Figure (b) depicts the changes in similarity between visual and textual features following SSAM application. All experiments are based on the ResNet-50 backbone.}
\label{fig:sm_alpha8beta}
\end{figure}

\begin{figure*}[t!]
\begin{center}
\includegraphics[width=1\linewidth]{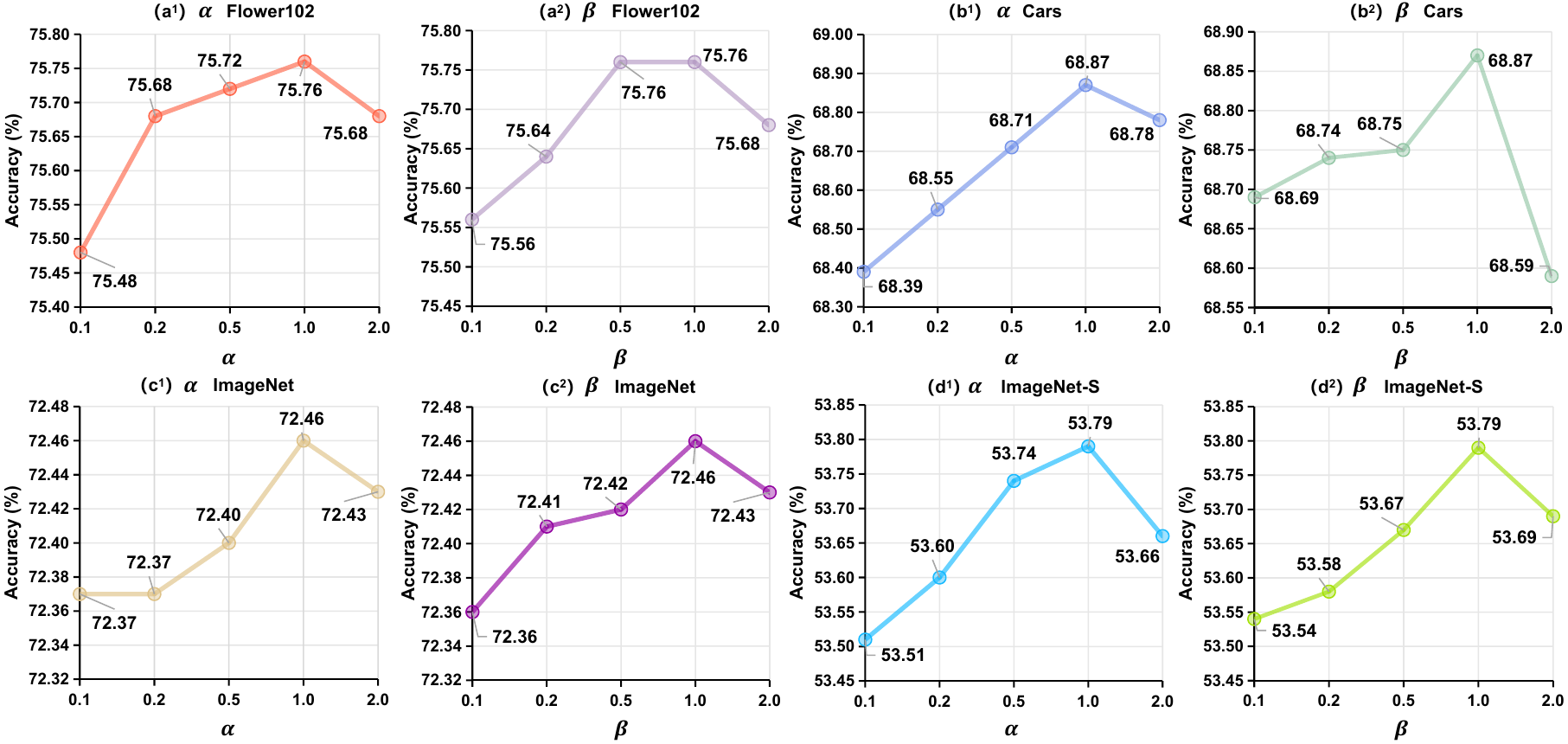}
\end{center}
\vspace{-0.4cm}
\caption{
\textbf{Results on typical datasets for ablation experiments on $\alpha$ and $\beta$}. Figures (a) and (b) show two representative datasets on the CD benchmark. Figures (c) and (d) show two representative datasets on the OOD benchmark.
}
\label{fig:sm_alpha8beta}
\end{figure*}

Additionally, to extend the visualization of the similarity between visual and textual features to the ResNet-50 backbone, we present supplementary visualization results in Figure~\ref{fig:sm_alpha8beta}~(b). From these results, it is evident that the visualization effects are consistent across both backbones. The similarity heatmap reveals a notable increase in the diagonal values after applying SSAM, further confirming the consistent effectiveness of our proposed method across different backbone networks and highlighting the advantages of SSAM in feature enhancement. This enhancement in feature similarity indicates that SSAM promotes a more coherent alignment between visual and textual features, facilitating improved cross-modal understanding and contributing to higher accuracy in image classification tasks through the integration of visual and textual information.



\section{Results on typical datasets for ablation experiments on $\alpha$ and $\beta$}
In the ablation experiments conducted on the hyper-parameters $\alpha$ and $\beta$, we perform five sets of experiments for each parameter, with each set covering 15 datasets. 
The primary focus is on evaluating the average performance across the CD and OOD benchmarks. In addition to the overall performance, individual datasets within both benchmark categories also yield highly favorable results.
For illustration, we have selected two representative datasets from each benchmark type: the Flower102 and Cars datasets from the CD benchmark, and ImageNet and ImageNet-S from the OOD benchmark. 
Detailed results are shown in Figure~\ref{fig:sm_alpha8beta}. The experimental results indicate that when both parameters $\alpha$ and $\beta$ are set to 1.0, the model achieves the best performance across all datasets. The optimization objective $\mathcal{L}_{CIR}$ facilitates unsupervised learning through image reconstruction, which significantly enhances the adaptability and stability of the visual encoder. Meanwhile, the optimization objective $\mathcal{L}_{ca}$  reinforces semantic consistency by associating visual features with textual features, thereby improving image classification accuracy. 
Experimental evidence demonstrates that increasing the weight of either $\alpha$ or $\beta$ individually diminishes the impact of the corresponding loss function, leading to a decline in model performance. Consequently, setting both $\alpha$ and $\beta$ to 1.0 ensures an optimal balance among the optimization objectives, resulting in the model's optimal performance.

\end{document}